\documentclass{article}
\usepackage{ijcai25}

\usepackage{times}
\usepackage{soul}
\usepackage{url}
\usepackage[hidelinks]{hyperref}
\usepackage[utf8]{inputenc}
\usepackage[small]{caption}
\usepackage{graphicx}
\usepackage{amsmath}
\usepackage{amsthm}
\usepackage{booktabs}
\usepackage{algorithm}
\usepackage{algorithmic}
\usepackage[switch]{lineno}

\usepackage{float}
\usepackage{multirow}
\usepackage{colortbl}
\definecolor{lightgray}{rgb}{0.9,0.9,0.9}
\usepackage{rotating}
\usepackage{amsthm,amsmath,amssymb}
\usepackage[toc,page]{appendix}
\usepackage{mathrsfs}
\usepackage{subcaption}

\usepackage{arydshln}

\title{Token-level Accept or Reject: A Micro Alignment Approach \\ 
for Large Language Models}

\author{
Yang Zhang$^{1}$
\and
Yu Yu$^{2}$\and
Bo Tang$^{3,2}$\footnote{Corresponding Author \textless tangbo@mail.ustc.edu.cn\textgreater.}\and
Yu Zhu$^4$\and
Chuxiong Sun$^5$\and
Wenqiang Wei$^{2}$\and\\
Jie Hu$^5$\and
Zipeng Xie$^6$\and
Zhiyu Li$^{2}$\and
Feiyu Xiong$^{2}$\And
Edward Chung$^1$\\
\affiliations
$^1$Hong Kong Polytechnic University, Hong Kong SAR, China\\
$^2$MemTensor (Shanghai) Technology Co., Ltd, Shanghai, China\\
$^3$University of Science and Technology of China, Suzhou Institute for Advanced Research, Suzhou, China\\
$^4$University of Science and Technology of China, Hefei, China\\
$^5$China Telecom Corporation Limited Beijing Research Institute, Beijing, China\\
$^6$Nanjing University of Information Science and Technology, Nanjing, China\\
\emails
}

\begin{document}

\maketitle


\begin{abstract} 
With the rapid development of Large Language Models (LLMs), aligning these models with human preferences and values is critical to ensuring ethical and safe applications. However, existing alignment techniques such as RLHF or DPO often require direct fine-tuning on LLMs with billions of parameters, resulting in substantial computational costs and inefficiencies. To address this, we propose \textbf{M}icro token-level \textbf{A}ccept-\textbf{R}eject \textbf{A}ligning (MARA) approach designed to operate independently of the language models. MARA simplifies the alignment process by decomposing sentence-level preference learning into token-level binary classification, where a compact three-layer fully-connected network  determines whether candidate tokens are “Accepted” or “Rejected” as part of the response. Extensive experiments across seven different LLMs and three open-source datasets show that MARA achieves significant improvements in alignment performance while reducing computational costs. The source code and implementation details are publicly available at \url{https://github.com/IAAR-Shanghai/MARA}, and the trained models are released at \url{https://huggingface.co/IAAR-Shanghai/MARA_AGENTS}.
\end{abstract}
\section{Introduction}

The alignment of Large Language Models (LLMs) with human values and preferences has emerged as a crucial challenge in AI development~\cite{wang2023aligning}. The alignment is extremely important for LLMs that operate safely and ethically. Among various alignment approaches, Reinforcement Learning from Human Feedback (RLHF)~\cite{ouyang2022training} and Direct Preference Optimization (DPO)~\cite{rafailov2024direct} have emerged as two dominant paradigms. RLHF fine-tunes language models using a reward model trained on human preference datasets, while DPO directly optimizes the models through pairwise comparisons without relying on explicit reward modeling.

Although RLHF, DPO and their variants~\cite{perez2022discovering,bai2022constitutional,lee2023rlaif,zeng2024token,azar2024general,wang2023beyond} have demonstrated impressive capabilities in aligning performance, they face a critical challenge: excessive computational resource consumption. These methods require fine-tuning billion or even hundred-billion parameter of language models, typically demanding hundreds of GPU hours and substantial memory resources. Such computational intensity poses a significant barrier for real-world applications~\cite{anwar2024foundational}, particularly in scenarios requiring rapid alignment updates or resource-constrained environments.

To address this challenge, we pose a fundamental question: \textit{Can we develop a micro alignment approach that operates independently of the language model while maintaining excellent alignment performance?}

\begin{figure}
\vspace{-1.5mm}	
\centering
\includegraphics[width=0.5\textwidth]{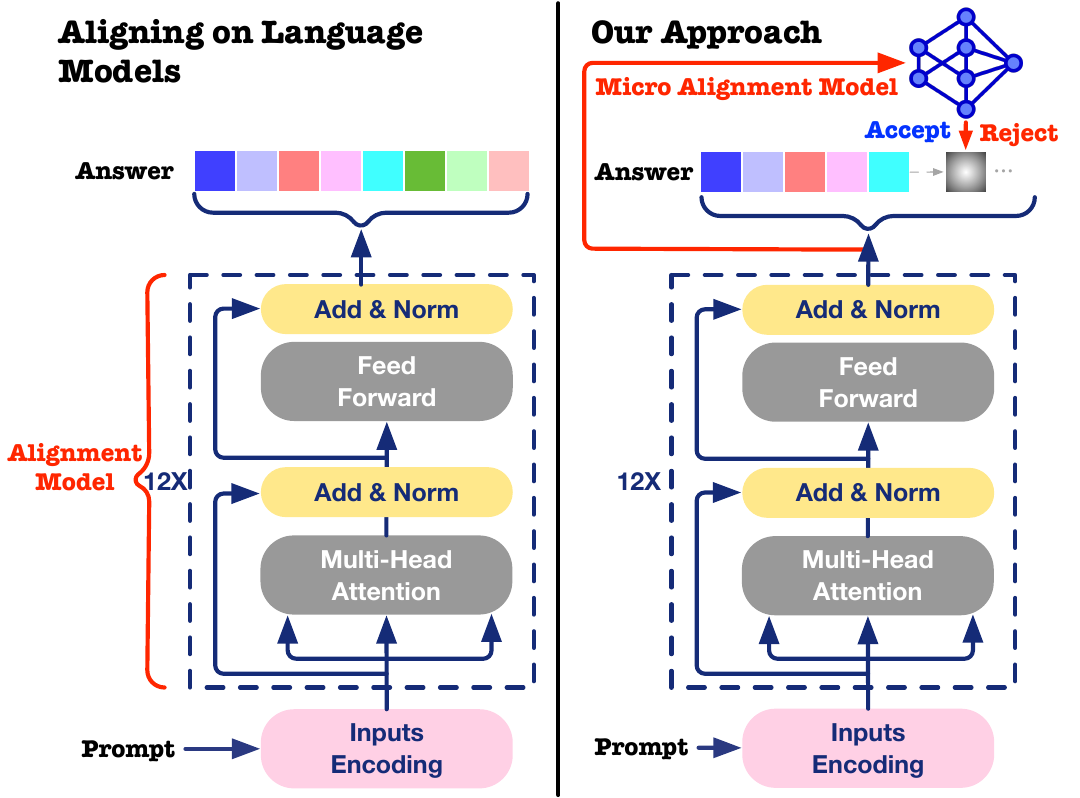}
\caption{Unlike RLHF or DPO based alignment approach which fine-tunes the language models, the key insight of our approach is simplifying the alignment process into accepting or rejecting a token using 
a streamlined alignment model.}
\vspace{-5mm}
\label{motivation}
\end{figure}

As shown in Figure~\ref{motivation}, unlike aligning operated on the language model, our key insight is that the alignment process can be simplified into accepting or rejecting a token using a streamlined alignment model. Specifically, the alignment model is the first to be implemented as a micro fully-connected network. Given the prompt, the partially generated response, and a candidate set of tokens that are sorted by sampling probabilities from a supervised fine-tuned model, our alignment model sequentially determines whether to accept or reject each candidate token during the generation process. This approach transforms the alignment task into a simple binary classification problem, significantly reducing the computational overhead while ensuring effective alignment performance.


A concurrent work called \textit{Aligner}~\cite{ji2024aligner} shares a similar motivation of decoupling alignment from the upstream language model. However, our approach differs significantly in both scale and methodology. While \textit{Aligner} employs a seq2seq model with 2B-13B parameters, our method utilizes a micro fully-connected network with merely millions of parameters. Moreover, our execution process is fundamentally simpler, focusing on token-level accept/reject decisions rather than reformulating entire output sentences.

In summary, we propose \textbf{M}icro token-level \textbf{A}ccept-\textbf{R}eject \textbf{A}ligning (MARA) approach, which offers three significant advantages:
\begin{itemize}
    \item Computation-friendly: \textbf{Our approach requires only a three-layer fully-connected network as the alignment model, reducing the computation overhead significantly.} Specifically, for aligning an 8B-parameter language model, MARA requires training only a 4M-parameter alignment model, whereas RLHF, DPO, and \textit{Aligner} necessitate training language models with over 20M parameters, even when employing parameter-efficient Low-Rank Adaptation (LoRA).
    \item Effectiveness: \textbf{While operating at a micro level, our approach achieves superior alignment performance}. Specifically, with Llama 3.1-8B as the base model, MARA demonstrates substantial improvements across different benchmark datasets: +31.8\% over RLHF, +18.8\% over DPO, and +8.8\% over \textit{Aligner}.
     \item Compatibility: The decoupled architecture of MARA-trained alignment models enables seamless integration with various LLMs, i.e., \textbf{an alignment model trained on one LLM can be effectively transferred to other LLMs while maintaining strong performance}. For instance, an alignment model trained on Mistral-7B-v0.3 enhances the alignment performance of Llama 3-8B by 25.5\% on average across three evaluation datasets.
\end{itemize}

We prioritize full reproducibility by releasing our complete implementation source code (available in the appendices). Our open-source commitment ensures that \textbf{all results presented in this paper can be independently verified and reproduced with minimal effort}, promoting transparency and facilitating future research in the field.




\section{Preliminaries}

This section introduces the standard RLHF framework, which consists of two primary phases: supervised fine-tuning (SFT) and reinforcement learning-based optimization. We first formalize these phases and then present a token-level decomposition of the alignment process.

\subsection{RLHF Framework}

\textbf{SFT Phase:} The pre-trained language model is initially fine-tuned on high-quality human demonstrations to generate appropriate responses. This process is optimized through the following objective:
\begin{equation}\label{eq:1}
J_{SFT} = - \mathbb{E}_{(x,y) \sim \mathcal{D}_{SFT}}\left[ {\log {\pi _{\mathrm{ref}} }\left( {y|x} \right)} \right]
\end{equation}
where $x$ and $y$ denote the input prompt and model response, respectively, sampled from the supervised fine-tuning dataset $\mathcal{D}_{SFT}$. $\pi _{\mathrm{ref}}$ denotes the fine-tuned language model serving as the reference model for subsequent optimization.

\noindent \textbf{RL-based Optimization Phase:} Following SFT, the model undergoes reinforcement learning optimization to align with human preferences. The objective function for this phase is:
\begin{equation}\label{eq:2}
\begin{aligned}
{J_{RL}} = & \mathbb{E}_{x \sim \mathcal{D}_{RL}, y \sim \pi_{\theta}\left( { \cdot |x} \right)}\left[ r\left( {x,y} \right) \right. \\
& \left. - \lambda {D_{KL}}\left( {{\pi _\theta }\left( { \cdot |x} \right)\parallel {\pi _{\mathrm{ref}}}\left( { \cdot |x} \right)} \right) \right]
\end{aligned}
\end{equation}
where $\pi_{\theta}$ denotes the model undergoing RL optimization, and $\mathcal{D}_{RL}$ represents the optimization dataset. The coefficient $\lambda$ controls the KL divergence penalty between the reference model and the aligned model.

The reward model $r(x,y)$ evaluates the alignment between model outputs and human preferences. Given a preference dataset $\mathcal{D}$ containing triples $(x,y_w, y_l)$, where $x$ is the prompt, $y_w$ is the preferred response, and $y_l$ is the less preferred response, the reward model is trained to minimize:
\begin{equation}\label{eq:3}
\mathcal{L}\left( {{r_\phi }} \right) =  - {\mathbb{E}_{\left( {x,{y_w},{y_l}} \right) \sim \mathcal{D}}}\left[ {\log \left( {\sigma \left( {{r_\phi }\left( {x,{y_w}} \right) - {r_\phi }\left( {x,{y_l}} \right)} \right)} \right)} \right]
\end{equation}
where $\sigma$ denotes the Sigmoid function.

\subsection{Token-level Decomposition}

At its core, the RLHF alignment process can be decomposed into a sequence of token-level decisions. The sentence generation process is formally expressed as:
\begin{equation}\label{eq:4}
{\pi _\theta }\left( {y|x} \right) = {\pi _\theta }\left( {{y_{1,}}{y_2}, \ldots ,{y_{{H}}}|x} \right) = \prod\limits_{i = 1}^{{H}} {\pi_{\theta}{{\left( {{y_i} |x,{y_{ < i}}} \right)}_{y_i \sim {{{\rm T}_i}}}}} 
\end{equation}
where ${{\rm T}}_i$ denotes the vocabulary space of available tokens at the $i$-th position in the output sequence $y$, and $H$ denotes the sequence length. 

This token-level decomposition reveals the possibility of performing alignment at the granularity of individual tokens, enabling more precise control over the generation process. Furthermore, this granular perspective can be refined into explicit accept-or-reject decisions for each token, which motivates our proposed method detailed in the following sections.

\section{Our approach}

\begin{figure*}[h]
\begin{center}
\includegraphics[width=1\textwidth]{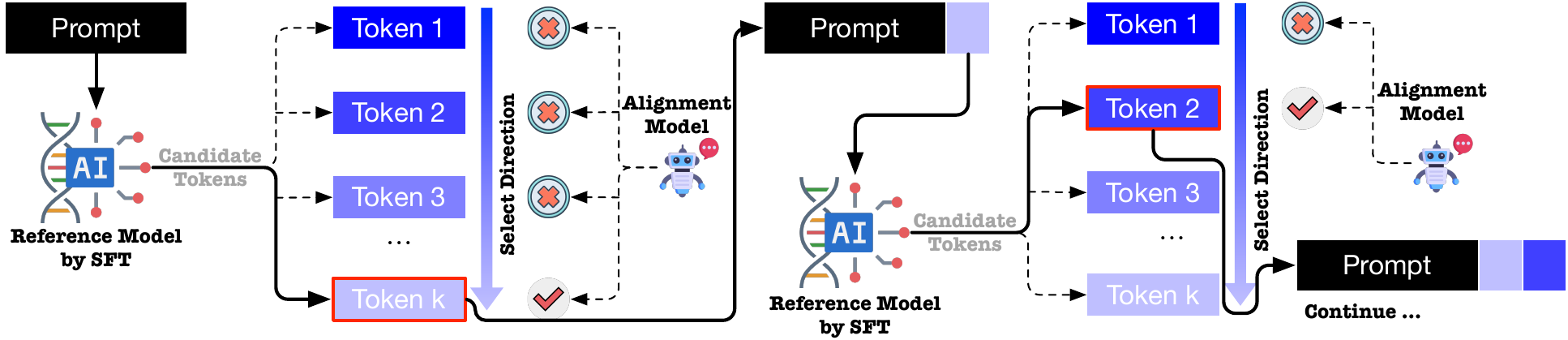}
\end{center}
\caption{Architecture of MARA: The alignment model performs token selection through accept-reject decisions.}
\vspace{-3mm}
\end{figure*}

In this section, we present MARA, a novel token-level accept-reject approach for LLM alignment. Building upon the foundation established by the SFT phase, MARA introduces a language model-independent alignment mechanism, replacing the original language model optimization in RLHF. Specifically, we first formalize the alignment task as a Markov Decision Process (MDP), then detail our accept-reject mechanism, and finally discuss implementation specifics.

\subsection{MDP formulation of alignment}

Given the reference model $\pi_{\mathrm{ref}}$ from the SFT phase, we formulate the alignment process as a MDP, characterized by the tuple $\left({\mathcal{S},\mathcal{A}, \rho, \mathcal{P},r} \right)$, where:

\begin{itemize}

    \item State space ($s \in \mathcal{S}$): For the $i$-th token generation at time step $\tau$, the state $s_{\tau} = \left\{ {{x},{y_{ < i}},{t_{i}^k}} \right\}$ comprises the input prompt $x$, previously generated response $y_{<i}$, and the current candidate token $t_{i}^k \in {\rm{\overline T }}_i$, where ${\rm{\overline T }}_i$ denotes a truncated candidate token set (detailed below), and $k$ indexes the candidate token in ${\rm{\overline T }}_i$.
    
    \item Action space ($a \in \{0,1\}$): Binary action where $a_{t_{i}^k} = 1$ indicates accepting the candidate token $t_{i}^k$ to concatenate with the generated response $y_{<i}$, while $a_{t_{i}^k} = 0$ indicates rejection. To facilitate understanding, we use $a_{t_{i}^k}$ instead of $a_{\tau}$ referring to the action on the candidate token.

    \item Initial state distribution ($\rho$): Defines the distribution over initial states.

    \item Reward function ($r: \mathcal{S} \times \mathcal{A} \rightarrow \mathbb{R}$) will be detailed in Section~\ref{details}.
    
    \item State transition ($\mathcal{P}$): The dynamics follow:
\end{itemize}
\begin{equation}\label{eqn:6} s_{\tau+1} = 
  \begin{cases}
      \left\{ {x,{y_{ < i + 1}},{t_{i + 1}^1}} \right\}, & \mbox{if } {a_{t_{i}^k} = 1};\\
      \left\{ {x,{y_{ < i}},{t_{i}^{k + 1}}} \right\}, & \mbox{else} .
  \end{cases}
\end{equation}

To enhance computational efficiency, we employ a hybrid vocabulary truncation strategy combining two widely-used methods: top-k sampling, which retains only the k most probable tokens, and top-p (nucleus) sampling, which selects the smallest set of tokens whose cumulative probability exceeds p. This combination clips the vocabulary space ${{\rm T}}_i$ at each generation step to form a reduced candidate set ${{\rm \overline T }}_i$. This hybrid approach ensures both diversity and quality of the candidate tokens while maintaining computational tractability. 

\subsection{Token-level Accept-Reject Aligning Mechanism}

Unlike traditional RLHF that fine-tunes the entire language model, our approach introduces a micro accept-reject model that operates at the token level. This model is implemented as a compact three-layer fully connected neural network. The alignment process follows four key steps:

\textbf{Step 1: Candidate Generation.}
Given the prompt $x$ and the previously generated tokens $y_{<i}$, we use the reference model $\pi_{\rm ref}$ to generate a candidate token set ${\rm{\overline T }}_i = \{t_{i}^1, t_{i}^2, \ldots, t_{i}^{|\rm{\overline{T} _i|}} \}$ through hybrid sampling that combines top-p and top-k strategies.

\textbf{Step 2: Probability-based Sorting.}
Sort the candidate token set ${\rm{\overline T }}_i$ in descending order according to their conditional probabilities $\pi_{\rm ref}(t_i^k|x,y_{<i})$, resulting in an ordered set ${\widetilde {\rm T}_i}$.

\textbf{Step 3: Token Evaluation.}
For the first token in the ${\widetilde {\rm T}_i}$, apply the accept-reject model to evaluate whether to accept or reject the token. The input of the accept-reject model consists of the last hidden state of the token sequence: prompt $x$, previously generated tokens $y_{<i}$, and the first candidate token in $\widetilde{T}_i$. The output is a binary decision: accept or reject. If accepting the token, add it to the output sentence $y$, and return to Step 1 until the \textbf{EOS} token is selected or the output length reaches its predefined limit. Otherwise, proceed to Step 4.

\textbf{Step 4: Candidate Update.}
Upon rejection, remove the token from the candidate token set ${\widetilde {\rm T}_i}$, and return to Step 3.

To maintain generation quality, we order candidate tokens by their sampling probabilities in descending order, as higher probability tokens typically better align with the reference model's learned patterns, thus reducing the risk of generating grammatically incorrect or contextually inconsistent content. 

For cases when all tokens in the candidate set are rejected by the accept-reject model, we employ a fallback mechanism to ensure generation continuity. Specifically, we enforce the acceptance of the final candidate token, formally expressed as $a_{t_{i}^k} \in \{1\}$ when ${k = \left| {\widetilde {\rm T}_i} \right|}$. This strategy ensures generation proceeds by selecting lower-probability tokens when necessary, thereby balancing quality and diversity.

According to the above token-level aligning approach, the sampling probability for any token $t_{i}^k$, representing the $k$-th candidate token in the candidate set ${\widetilde {\rm T}_i}$, is formalized as:
\begin{equation}\label{eq:6}
\begin{aligned}
\pi_{\theta}\left( {y_i = t_{i}^k} \right) = & \pi_{\theta}\left( a_{t_{i}^k}=1|x,y_{<i} \right)\\
& \prod\limits_{k' = 1}^{k - 1} \pi_{\theta}\left( a_{t_{i}^{k'}}=0|x,y_{<i} \right) \bigg|_{t_{i}^k \in \widetilde{\rm {T}}_i}.
\end{aligned}
\end{equation}
Note that we still use $\pi_{\theta}$ to denote the accept-reject model parameterized by $\theta$.







From above, the complete aligned response generation process can thus be expressed as:
\begin{equation}\label{eq:Proxy-pi}
{\pi _\theta }\left( {y|x} \right) = \prod\limits_{i = 1}^{H} {\pi_{\theta}({{y_i} = {t_{i}^k}|x,{y_{ < i}}} )}\big|_{t \sim {\widetilde {\rm T}_i}}.
\end{equation}

\subsection{Implementation}\label{details}


\subsubsection{Reward Design}

A key challenge in MARA is bridging the gap between sentence-level rewards in standard RLHF and token-level decisions. Our reward function is designed with two objectives: (1) maximizing the reward model score $r(x,y)$ to maintain alignment with human preferences as in RLHF, and (2) preventing reward hacking by constraining the token distribution divergence between the aligned model and the reference model, thereby avoiding unnatural token selections~\cite{eisenstein2023helping}.

Specifically, for any action $a_{t_i^k}$ at state $s_{\tau}$, we define the token-level reward as:
\begin{equation}\label{eqn:8} 
r_{\tau} = 
\begin{cases}
    -\lambda D_{KL}(\pi_{\theta}(t_{i}^k|x,y_{<i}) \parallel 
    \pi_{\text{ref}}(t_{i}^k|x,y_{<i})), \\
    \quad \text{if $s_{\tau}$ is not terminal};\\[8pt]
    r(x,y) - \lambda D_{KL}(\pi_{\theta}(t_{i}^k|x,y_{<i}) \parallel 
    \pi_{\text{ref}}(t_{i}^k|x,y_{<i})), \\
    \quad \text{else}.
\end{cases}
\end{equation}
where $r(x,y)$ denotes the evaluation score by reward model trained by Eq.~\ref{eq:3}.

This formulation represents a token-level decomposition of the RLHF objective in Eq.~\ref{eq:2}. We adopt the distributed Soft Actor-Critic (SAC) algorithm~\cite{haarnoja2018soft} instead of the conventional Proximal Policy Optimization (PPO) algorithm used in RLHF due to SAC's higher computation efficiency, especially in distributed training environment. PPO requires synchronization between distributed nodes for sample collection and model updates, creating waiting periods. In contrast, SAC enables continuous sample collection across all nodes with real-time parameter updates via a global replay buffer, eliminating waiting periods.

\subsubsection{Training Objectives}

The alignment model serves as the actor in our framework, with its loss function defined as:
\begin{equation}\label{eqn:9} 
\begin{aligned}
\mathcal{L}\left( {{\pi _\theta }} \right)  = - & {\mathbb{E}_{x \sim \mathcal{D}}}[ {\alpha_h}\mathcal{H}\left( {{\pi _\theta }\left( s \right)} \right) \\
&  + {\mathbb{E}_{t_k^i \sim {\pi _\theta }}}\left[ {\min \left( {{V_1}\left( s \right),{V_2}\left( s \right)} \right)} \right]]
\end{aligned}
\end{equation}
where $\mathcal{H}$ represents the entropy term of the alignment model that promotes diverse token selection, while $V_1(s)$ and $V_2(s)$ are two critic heads for robust value estimation and overfitting prevention.

The critic network, parameterized by $\varphi$, is trained using the following loss:
\begin{equation}\label{eqn:10}
\begin{aligned}
\mathcal{L}(\pi_{\varphi}) = \mathbb{E}_{x \sim \mathcal{D}} & \Bigg[ \bigg(\frac{V_1(s) + V_2(s)}{2} - r(x,\pi_{\theta}(x)) \\
& - \alpha_h \mathcal{H}(\pi_{\theta}(s')) \\
& - \gamma \mathbb{E}_{t_{k}^{i'} \sim \pi_{\theta}} \big[\min(V_1'(s'),V_2'(s'))\big]\bigg)^2 \Bigg]
\end{aligned}
\end{equation}
where $\pi_{\varphi}$ denotes the critic network with parameters ${\varphi}$. $V_1'(s')$ and $V_2'(s')$ denote the two heads of target critic network.

The entropy coefficient $\alpha_h$ is automatically adjusted according to:
\begin{equation}\label{eqn:11}
\begin{aligned}
\mathcal{L}(\alpha_h) = \mathbb{E}_{a_{h} \sim \pi_{\theta}} \left[-\alpha_{h}\pi_{\theta}\left(a_{\tau}|s_{\tau}\right) - \alpha_{\tau} \mathcal{\overline H }\right]
\end{aligned}
\end{equation}
where $\mathcal{\overline H}=2$ is the minimum entropy threshold based on the action space dimension.

\section{Experiments}

We conduct comprehensive experiments to evaluate MARA's effectiveness across diverse datasets, reward models, and base LLMs. 


\subsection{Experimental Setup}

\noindent \textbf{Datasets.} Our experiments leverage three comprehensive evaluation benchmarks for utility, safety and ethical assessment: PKU-SafeRLHF (SafeRLHF)~\cite{ji2024pku}, which serves as our primary training dataset and contains 83.4K preference pairs with safety meta-labels and human preferences for helpfulness and harmlessness; BeaverTails~\cite{ji2024beavertails}, covering 14 safety categories including sensitive topics like abuse and political discussions; and HarmfulQA~\cite{bhardwaj2023red}, comprising 10 themes of ChatGPT-generated conversations for evaluating responses in potentially harmful scenarios.

\noindent \textbf{Upstream LLMs.} We evaluate our alignment model on two widely-adopted open-source LLM families with various model scales and versions. Specifically, the Llama family includes Llama-3-8B, Llama-3.1-8B, Llama-3.2-3B, and Llama-3.2-1B~\cite{llama3modelcard}, while the Mistral family comprises Mistral-7B-v0.1, Mistral-7B-v0.2, and Mistral-7B-v0.3~\cite{jiang2023mistral}. Due to computational constraints, we do not include larger models such as Llama-3-70B in our experiments. For brevity, we omit the suffix 'Instruct' from all model names.

\noindent \textbf{Evaluation Metrics.} Following the methodology of \textit{aligner}~\cite{ji2024aligner}, we adopt the preference rate metric to evaluate model performance across two critical dimensions: helpfulness and harmlessness, which respectively assess the utility and safety aspects of generated responses. The preference rate is defined as:
\begin{equation}
w = \frac{{N_w} - {N_l}}{{N_w} + {N_e} + {N_l}} \times 100\% 
\end{equation}
where ${N_w}$, ${N_e}$, and ${N_l}$ denote the numbers of wins, ties, and loses respectively in pairwise comparisons between different alignment approaches. A comprehensive description of the evaluation metrics is provided in Appendix\ref{app:metrics}.

\noindent \textbf{Reward Models.} Our alignment training utilizes a combination of two specialized models: beaver-7b-v1.0-reward for utility evaluation and beaver-7b-v1.0-cost~\cite{dai2023safe} for safety evaluation (where a negative cost score indicates a safe response). The final reward $r(x,y)$ is computed as:
\begin{equation}
r(x,y)= \alpha_{r}R(x,y) - \alpha_{c}C(x,y)
\end{equation}
where $R(x,y)$ and $C(x,y)$ denote the evaluation scores from the reward and cost models respectively, and $\alpha_{r}$ and $\alpha_{c}$ are their corresponding weights.

For the Llama family of models, we employ a balanced weighting scheme with $\alpha_{r}=\alpha_{c}=1$. For the Mistral family, we adopt an asymmetric weighting ($\alpha_{r}=2$, $\alpha_{c}=1$) to counteract their inherent bias towards safety over utility, thereby achieving a better utility-safety trade-off.


\noindent \textbf{Computing Resources.} Our experiments are performed on one Nvidia H800-80GB GPU. The machine is equipped with 192 Intel(R) Xeon(R) Platinum 8468v processors and has a CPU memory of 1584 GB.

\subsection{Experiment Results}

\noindent \textbf{Performance on Different Evaluation Datasets.}

\begin{table}[t]
\centering
\caption{Performance improvements of MARA across PKU-SafeRLHF, BeaverTails, and HarmfulQA datasets. Each entry shows the percentage improvement in preference rate achieved by applying MARA compared to using the original LLM alone.}
\scalebox {0.9}{
\begin{tabular}{lrrr}
\toprule
& \multicolumn{3}{c}{Upstream LLM + MARA \textit{vs.} Upstream LLM} \\ \cmidrule(lr){2-4}
Upstream LLM &  SafeRLHF & BeaverTails & HarmfulQA \\
\midrule
Llama 3-8B & +33.67\% & +36.86\% & +52.05\%  \\
Llama 3.1-8B & +39.70\% & +35.43\% & +62.09\%\\
Llama 3.2-1B & +45.23\% & +31.71\% & +59.43\%  \\
Llama 3.2-3B & +32.66\% & +29.43\% & +35.45\%  \\
Mistral-7B-v0.1 & +11.06\% & +17.43\% & +10.45\%  \\
Mistral-7B-v0.2 & +4.52\% & +14.29\% & +4.92\%  \\
Mistral-7B-v0.3 & +17.09\% & +16.00\% & +11.07\% \\
\rowcolor{lightgray}
Average & +26.28\% & +25.88\% & +33.64\% \\
\bottomrule
\end{tabular}}
\label{tab:main_1}
\vspace{-3mm}
\end{table}

Table~\ref{tab:main_1} shows that MARA significantly improves the alignment performance of various upstream LLMs with different scales and versions in all three challenging datasets. For the Llama family, MARA achieves remarkable improvements, with gains of up to +45.23\% on SafeRLHF, +36.86\% on BeaverTails, and +62.09\% on HarmfulQA. The improvements are particularly pronounced for Llama 3.1-8B, which shows the strongest overall performance gains. For the Mistral family, while the improvements are more modest, MARA still demonstrates consistent positive impacts, with average gains of +10.89\%, +15.91\%, and +8.81\% across the three datasets respectively. On average, MARA yields substantial improvements of +26.28\%, +25.88\%, and +33.64\% across SafeRLHF, BeaverTails, and HarmfulQA respectively, indicating its robust generalization capability across different evaluation scenarios. More detailed experimental results, including the win/tie/lose statistics across different models and datasets, are presented in Appendix~\ref{app:details}.

\noindent \textbf{Comparison with Different Baselines.}

\begin{table*}[t]
\centering
\caption{Performance comparison of MARA against RLHF, DPO, and \textit{Aligner}  measured by percentage improvements of preference rate.}
\scalebox {0.84}{
\begin{tabular}{lrrrrrrrrr}
\toprule
&\multicolumn{3}{c}{MARA \textit{vs.} RLHF} & \multicolumn{3}{c}{MARA \textit{vs.} DPO} & \multicolumn{3}{c}{MARA \textit{vs.} \textit{Aligner}}\\
\cmidrule(lrr){2-4} \cmidrule(lrr){5-7} \cmidrule(lrr){8-10}
Upstream LLM & SafeRLHF & BeaverTails & HarmfulQA & SafeRLHF & BeaverTails & HarmfulQA & SafeRLHF & BeaverTails & HarmfulQA\\
\midrule
Llama 3-8B & +29.65\% & +22.14\% & +45.08\% & +15.08\% & +14.86\% & +28.28\% & +5.53\% & +2.71\% & +5.74\%\\
Llama 3.1-8B & +21.61\% & +28.00\% & +5.74\% & +0.00\% & +14.57\% & +1.64\% & +0.50\% & -3.14\% & -11.89\%\\
Llama 3.2-1B & +39.20\% & +22.29\% & +38.93\% & +0.50\% & +0.14\% & +14.96\% & -6.03\% & -12.0\% & -9.63\%\\
Llama 3.2-3B & +8.54\% & +7.43\% & -1.02\% & +1.01\% & +2.71\% & -4.92\% & -2.01\% & -6.71\% & -12.91\%\\
Mistral-7B-v0.1 & +8.04\% & -0.14\% & +0.00\% & +9.55\% & +17.29\% & +6.15\% & +10.55\% & +17.0\% & +0.61\%\\
Mistral-7B-v0.2 & +24.62\% & +16.86\% & +17.01\% & +22.61\% & +27.0\% & +21.93\% & +14.57\% & +13.43\% & +4.30\%\\
Mistral-7B-v0.3 & +14.07\% & +16.86\% & +26.64\% & +17.09\% & +29.86\% & +23.98\% & +16.08\% & +19.29\% & +11.48\%\\
\rowcolor{lightgray}
Average & +20.82\% & +16.21\% & +18.91\% & +9.41\% & +15.20\% & +13.15\% & +5.60\% & +4.37\% & -1.76\%\\
\bottomrule
\end{tabular}}
\label{tab:main_2}
\vspace{-3mm}
\end{table*}

Table~\ref{tab:main_2} presents a comprehensive comparison between MARA and three representative baselines: RLHF, DPO, and \textit{Aligner}. For the implementation of RLHF and DPO, we utilize the source code from \cite{zheng2024llamafactory}; For the implementation of \textit{Aligner}, we utilize the aligner-7b-v1.0 alignment model from~\cite{huggingfaceAligner}.
Note that we don't include recent variants of RLHF and DPO (e.g., RLAIF, RLHAIF, TDPO, IPO) as baselines since they mainly improve specific aspects (e.g., feedback sources, overfitting) rather than proposing fundamentally different alignment mechanisms. The experimental results demonstrate that MARA consistently outperforms RLHF and DPO across all three datasets, achieving average improvements of +18.65\% and +12.59\% respectively. 

Compared to \textit{Aligner}, MARA shows marginally lower performance on certain Llama-based models, likely because \textit{Aligner} was trained using Llama models as base models. Nevertheless, MARA demonstrates competitive advantages on SafeRLHF (+5.60\%) and BeaverTails (+4.37\%) evaluation datasets, while showing stronger ablation effects on HH-RLHF and Ultra-Feedback datasets (Table~\ref{tab:abla_data}). More importantly, MARA shows significant computational advantages: it requires only a micro three-layer alignment model (4M parameters) compared to \textit{Aligner}'s 7B parameters, achieving higher inference speed (31.41 \textit{vs.} 20.63 tokens/s).


\begin{table}[t]
\centering
\setlength{\tabcolsep}{1.7pt}
\caption{Compatibility analysis for our approach, that an alignment model trained with a LLM to be aggregate with other inference LLM. The value of each cell represents the percentage improvement in preference rate of our algorithm over the upstream model, \textit{i.e.,} inference model.}
\scalebox {0.86}{
\begin{tabular}{llrrr}
\toprule
& & \multicolumn{3}{c}{MARA \textit{vs.} Inference LLM} \\
\cmidrule(lr){3-5}
Training LLM & Inference LLM & SafeRLHF & BeaverTails & HarmfulQA \\
\midrule
Llama-3.1-8B & Llama 3.1-8B & +39.70\% & +35.43\% & +62.09\% \\
& Llama 3-8B & +27.14\% & +28.14\% & +45.29\% \\
& Mistral-7B-v0.3 & +14.07\% & +5.00\% & +6.15\% \\
& Mistral-7B-v0.2 & +4.02\% & +7.00\% & +7.17\% \\
& Mistral-7B-v0.1 & +17.09\% & +12.57\% & +1.64\% \\
\cmidrule[0.5pt]{1-5}
Mistral-7B-v0.3 & Mistral-7B-v0.3 & +9.05\% & +9.14\% & +6.97\% \\
& Mistral-7B-v0.2 & +3.02\% & +1.71\% & +1.43\% \\
& Mistral-7B-v0.1 & +4.52\% & +6.86\% & +4.30\% \\
& Llama 3.1-8B & +1.51\% & +12.57\% & +5.94\% \\
& Llama 3-8B & +18.09\% & +20.71\% & +37.70\% \\
\cmidrule[0.5pt]{1-5}
\rowcolor{lightgray}
Average & & +13.82\% & +13.91\% & +17.87\% \\
\bottomrule
\end{tabular}}
\label{tab:compat}
\vspace{-2mm}
\end{table}




\noindent \textbf{Compatibility Analysis.}

To verify the compatibility of our approach, we conduct comprehensive experiments examining how alignment models trained on one LLM generalize to other inference LLMs. Table~\ref{tab:compat} presents the results for two alignment models, trained on Llama-3.1-8B and Mistral-7B-v0.3 respectively, when applied to various inference LLMs. The results show that the alignment model trained on Llama-3.1-8B demonstrates strong generalization capability, achieving substantial improvements across all evaluation datasets. Even when applied to different model families, it maintains robust performance, with improvements ranging from +4.02\% to +28.14\% across different Mistral versions. Similarly, the Mistral-7B-v0.3-trained alignment model shows consistent improvements across both model families, with particularly strong performance when applied to Llama 3-8B (SafeRLHF: +18.09\%, BeaverTails: +20.71\%, HarmfulQA: +37.70\%). Overall, our approach demonstrates strong cross-model compatibility with average improvements of +13.82\%, +13.91\%, and +17.87\% across SafeRLHF, BeaverTails, and HarmfulQA respectively.

\subsection{Ablation Studies}\label{ablation}

\noindent \textbf{Ablation on Training Datasets}

\begin{table}[t]
\centering
\caption{Performance comparison of MARA against baseline approaches (SFT, RLHF, DPO, and \textit{Aligner}) on various datasets, reported as percentage improvements in preference rate. All experiments use Llama 3.1-8B as the upstream model.}
\scalebox{0.88}{
\begin{tabular}{lrrrr}
\toprule
Dataset & {\textit{vs.} SFT} & {\textit{vs.} RLHF} & {\textit{vs.} DPO} & {\textit{vs.} \textit{Aligner}}\\
\midrule
HH-RLHF & +24.00\% & +43.00\% & +28.50\% & +8.50\% \\
Ultra-Feedback & +22.50\% & +20.50\% & +9.00\% & +9.00\% \\
\rowcolor{lightgray}
Average & +23.25\% & +31.75\% & +18.75\% & +8.75\% \\
\bottomrule
\end{tabular}}
\label{tab:abla_data}
\vspace{-4mm}
\end{table}

The experiments introduced before utilized PKU-SafeRLHF datasets for training the alignment model. To verify MARA's effectiveness across different training datasets, we evaluate its performance against SFT, RLHF, DPO and \textit{Aligner} on two additional datasets: HH-RLHF~\cite{bai2022training} and Ultra-Feedback~\cite{cui2023ultrafeedback}. Table~\ref{tab:abla_data} presents the results using Llama 3.1-8B as the upstream LLM. The experimental results demonstrate MARA's consistent superiority, with substantial average improvements over all baselines: +23.25\% over SFT, +31.75\% over RLHF, +18.75\% over DPO, and +8.75\% over \textit{Aligner}. Notably, MARA achieves particularly strong performance on HH-RLHF, with improvements of up to +43.00\% over RLHF, while maintaining robust gains on Ultra-Feedback across all baseline comparisons.

\noindent \textbf{Ablation on the Reward Signal Distribution} 

\begin{table*}[t]
\centering
\setlength{\tabcolsep}{2pt}
\scalebox{0.83}{
\begin{tabular}{llrrrrrrrrr}
\toprule
& \multirow{2}{*}[-1.1ex]{\begin{tabular}[c]{@{}l@{}}Reward \\ distribution\end{tabular}} & \multicolumn{3}{c}{SafeRLHF} & \multicolumn{3}{c}{BeaverTails} & \multicolumn{3}{c}{HarmfulQA}\\ \cmidrule(lrr){3-5} \cmidrule(lrr){6-8} \cmidrule(lrr){9-11}
Upstream LLM & & {Helpful($\uparrow$)} & {Harmless($\uparrow$)} & {Perference($\uparrow$)} & {Helpful($\uparrow$)} & {Harmless($\uparrow$)} & {Perference($\uparrow$)} & {Helpful($\uparrow$)} & {Harmless($\uparrow$)} & {Perference($\uparrow$)} \\
\midrule
\multirow{4}{*}{Mistral-7B-v0.1}
& $\alpha_r$:$\alpha_c$=1:0 & \textbf{+29.65\%} & -26.13\% & +2.01\% & \textbf{+37.29\%} & -22.57\% & +7.29\% & \textbf{+21.93\%} & -18.44\% & +1.43\% \\
& $\alpha_r$:$\alpha_c$=0:1 & -55.78\% & \textbf{+63.82\%} & +4.02\% & -34.29\% & \textbf{+51.57\%} & +8.57\% & -49.39\% & \textbf{+53.89\%} & +2.46\% \\
& $\alpha_r$:$\alpha_c$=1:1 & -37.49\% & +49.25\% & +6.03\% & -19.71\% & +48.71\% & +14.43\% & -40.78\% & +51.84\% & +5.53\% \\
& $\alpha_r$:$\alpha_c$=2:1 & -16.58\% & +28.69\% & \textbf{+11.06\%} & -2.71\% & +37.21\% & \textbf{+17.43\%} & -22.75\% & +43.44\% & \textbf{+10.45\%} \\
\midrule
\multirow{4}{*}{Mistral-7B-v0.2}
& $\alpha_r$:$\alpha_c$=1:0 & \textbf{+24.62\%} & -20.60\% & +2.01\% & \textbf{+31.71\%} & -21.14\% & +5.29\% & \textbf{+27.05\%} & -22.95\% & +2.05\% \\
& $\alpha_r$:$\alpha_c$=0:1 & -91.96\% & \textbf{+40.70\%} & -1.01\% & -30.00\% & +33.71\% & +1.86\% & -52.87\% & \textbf{+52.46\%} & -0.20\% \\
& $\alpha_r$:$\alpha_c$=1:1 & -22.61\% & +35.68\% & \textbf{+6.53\%} & -15.43\% & \textbf{+35.14\%} & +9.71\% & -41.19\% & +50.41\% & +4.51\% \\
& $\alpha_r$:$\alpha_c$=2:1 & -20.6\% & +30.15\% & +4.52\% & +9.43\% & +19.29\% & \textbf{+14.29\%} & -31.56\% & +41.6\% & \textbf{+4.92\%} \\
\midrule
\multirow{4}{*}{Mistral-7B-v0.3}
& $\alpha_r$:$\alpha_c$=1:0 & \textbf{+29.65\%} & -9.05\% & +10.55\% & \textbf{+37.57\%} & -11.29\% & +13.00\% & \textbf{+11.07\%} & -9.84\% & +0.61\% \\
& $\alpha_r$:$\alpha_c$=0:1 & -39.70\% & \textbf{+65.83\%} & +13.07\% & -37.71\% & \textbf{+45.71\%} & +4.00\% & -56.97\% & +60.25\% & +1.64\% \\
& $\alpha_r$:$\alpha_c$=1:1 & -31.66\% & +54.27\% & +11.56\% & -17.14\% & +43.14\% & +13.00\% & -50.41\% & \textbf{+69.06\%} & +9.22\% \\
& $\alpha_r$:$\alpha_c$=2:1 & -17.39\% & +50.75\% & \textbf{+17.09\%} & -2.29\% & +34.71\% & \textbf{+16.00\%} & -42.62\% & +64.75\% & \textbf{+11.07\%} \\
\bottomrule
\end{tabular}}
\caption{Impact of reward signal distribution on model performance in terms of helpful, harmless, and preference rate. $\alpha_r$:$\alpha_c$ represents the ratio between reward model (beaver-7b-v1.0-reward) and cost model (beaver-7b-v1.0-cost) signals in the reward of $r(x,y)$. \textbf{Bold} numbers indicate the best performance under each metric (helpful/harmless/preference) for each model version.}
\label{tab:abl_distri}
\vspace{-4mm}
\end{table*}

To analyze the impact of different reward signal distributions, we conduct experiments with various ratios between reward and cost model signals ($\alpha_r$:$\alpha_c$). As shown in Table~\ref{tab:abl_distri}, we observe distinct trade-offs between helpfulness and harmlessness across different configurations. When solely using the reward model ($\alpha_r$:$\alpha_c$=1:0), the model shows improved helpfulness (up to +37.57\%) but decreased harmlessness. Conversely, using only the cost model ($\alpha_r$:$\alpha_c$=0:1) leads to significant gains in harmlessness (up to +65.83\%) but often at the expense of helpfulness. A balanced ratio of $\alpha_r$:$\alpha_c$=2:1 achieves the best overall performance across datasets, with preference rate improvements reaching +17.09\% on SafeRLHF, +16.00\% on BeaverTails, and +11.07\% on HarmfulQA using Mistral-7B-v0.3. This suggests that while both reward signals are important, slightly emphasizing the reward model over the cost model leads to optimal balance between helpfulness and safety. Detailed ablation studies on the impact of candidate token set size and the sensitivity to KL divergence coefficients are presented in Appendix~\ref{app:size} and Appendix~\ref{app:KL}, respectively.

\subsection{Visualisation of Accepted Tokens Distribution}

\begin{figure}[t]
    \centering
    \includegraphics[width=0.47\textwidth]{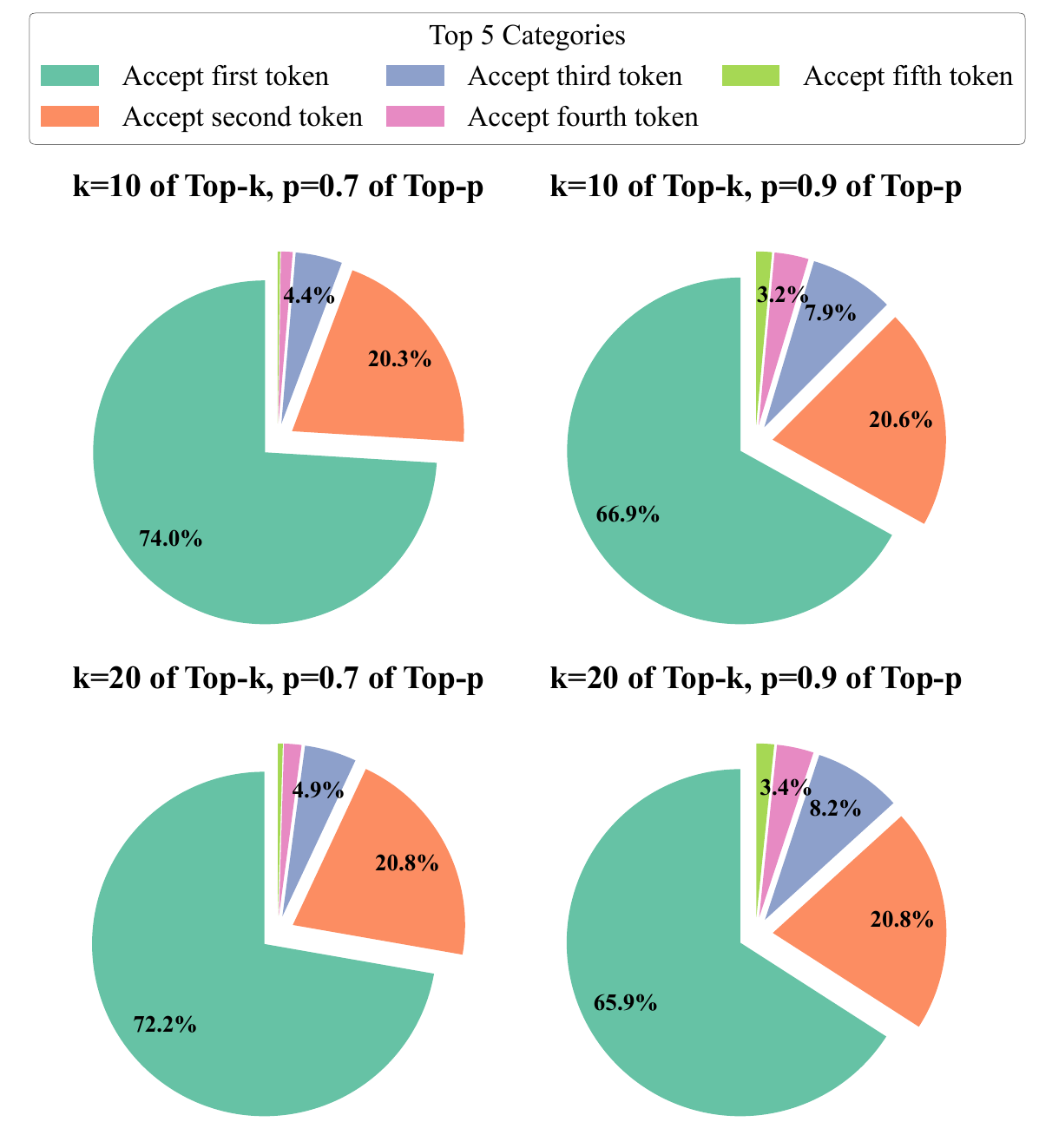}
    \caption{Token acceptance distribution under various Top-k and Top-p sampling configurations using Llama 3.2-1B on the PKU-SafeRLHF dataset. The plots show the first five most frequently accepted token positions, with acceptance rates below 2\% omitted.}
    \label{fig:token_distribution}
    \vspace{-4mm}
\end{figure}

To visualize the accept or rejection process for candidate tokens under our approach, and evaluate the impact of sampling parameters on token selection, we present an instance of the experiment performed on PKU-SafeRLHF dataset with Llama-3.2-1B as the upstream LLM. More experiments are provided in the Appendix~\ref{app:vis}.

Figure~\ref{fig:token_distribution} shows the distribution of accepted tokens across the candidate token set under different sampling parameter configurations. The analysis reveals three key findings:

1) Dominance of first token: Across all parameter settings, the first candidate token demonstrates strong dominance, accounting for 65.9-74.0\% of all acceptances. This suggests the model maintains high confidence in its primary predictions with the greatest sampling probability.

2) Parameter sensitivity:
Increasing the Top-p value and Top-k value results in a notable decrease in first token acceptance, while simultaneously increasing the acceptance rates of second and third tokens. This indicates that higher sampling thresholds promote more diverse token selection.

3) Optimal decision window: The acceptance distribution demonstrates MARA has learned to concentrate its safety-alignment decisions within a compact token space, where over 95\% of all acceptances occur within the top three positions. Such a focused decision mechanism helps maintain generation efficiency while ensuring safety controls.

\section{Related Works}
Prior approaches to language model alignment can be categorized along two primary dimensions: the utilization of RL and the granularity of alignment.

\subsection{Alignment Methodology}

\textbf{RL-based Alignment:} RL has demonstrated remarkable effectiveness across various language tasks, including question-answering, machine translation, and text summarization~\cite{kang2020dynamic,ziegler2019fine,stiennon2020learning,nakano2021webgpt}. Recent years have witnessed its successful application in aligning language models with human values and preferences. State-of-the-art language models, notably InstructGPT~\cite{ouyang2022training} and Llama2~\cite{touvron2023llama}, leverage RL-based fine-tuning. A prominent paradigm in this domain is RLHF, which constructs reward models from human preference data to guide model fine-tuning. Building upon RLHF, RLxF extends this framework to incorporate diverse feedback sources beyond human responses. The variable $x$ in RLxF encompasses AI feedback (RLAIF)~\cite{bai2022constitutional,lee2023rlaif} and hybrid human-AI feedback (RLHAIF)~\cite{wu2021recursively,saunders2022self,perez2022discovering}.

\noindent \textbf{Non-RL Alignment:} Given the implementation complexity and computational demands of RL-based approaches, alternative methodologies have emerged. RAFT~\cite{dongraft} and RRHF~\cite{yuan2023rrhf} employ selective fine-tuning on high-quality samples. Rain~\cite{lirain} and \textit{Aligner}~\cite{ji2024aligner} adopt output rectification strategies, with Rain utilizing an evaluation-rewind mechanism and \textit{Aligner} implementing a supervised, decoupled alignment model. DPO~\cite{rafailov2024direct} establishes a theoretical mapping between optimal policies and reward functions, eliminating explicit reward modeling. Recent extensions—IPO~\cite{azar2024general}, Token-DPO~\cite{zeng2024token}, and DPO-f~\cite{wang2023beyond}—address challenges in overfitting, fine-grained alignment, and reverse KL regularization constraints, respectively. Our approach maintains alignment quality comparable to RL-based methods while achieving computational efficiency superior to existing non-RL techniques.

\subsection{Alignment Granularity}

\textbf{Sentence-Level Alignment:} In most alignment research, the alignment environment is modeled as a bandit environment, treating complete sentences as atomic actions. This paradigm encompasses RLHF~\cite{ouyang2022training} and its variants (RLAIF~\cite{bai2022constitutional,lee2023rlaif}, RLHAIF~\cite{wu2021recursively,saunders2022self,perez2022discovering}), as well as DPO-based methods~\cite{rafailov2024direct,azar2024general,wang2023beyond}. Additional approaches in this category include Rain and \textit{Aligner}'s comprehensive sentence correction mechanisms, and RAFT and RRHF's high-quality sentence filtering strategies.

\noindent \textbf{Token-Level Alignment:} Recently, there has been increased interest in token-level alignment, decomposing the task into sequential token generation decisions for enhanced granularity and control. Both RLHF~\cite{zhong2024dpo} and DPO~\cite{zeng2024token,rafailov2024r} have evolved to support token-level alignment through MDP formulations, enabling more precise intervention at each generation step. MARA uniquely decouples the alignment model from the language model while operating at the token level. This novel approach significantly reduces computational costs compared to existing coupled alignment methods, making MARA the first efficient decoupled token-level alignment framework.

\section{Conclusion}

This paper proposes MARA, a micro alignment approach that enhances LLMs' adherence to human preferences through token-level control. Our key innovation lies in introducing a micro alignment model that operates independently from the base language model, making \textbf{Accept/Reject} decisions for candidate tokens to achieve fine-grained alignment. Implemented as a compact three-layer fully-connected network, MARA significantly reduces computational overhead compared to existing SOTA approaches while maintaining superior alignment performance. Extensive experiments across multiple evaluation datasets and LLM architectures demonstrate MARA's effectiveness.

\section*{Ethics Statement}

Our research on MARA presents a novel approach to language model alignment that significantly reduces computational requirements while maintaining alignment effectiveness. Our experimental validation utilizes publicly available datasets like PKU-SafeRLHF, BeaverTails and HarmfulQA, which are designed to promote helpful, harmless, and honest AI responses. While our efficient alignment approach could theoretically be misused for harmful purposes, we strongly oppose any malicious applications and advocate for the responsible development of alignment techniques.

To promote transparency and reproducibility, we release our complete implementation code under open-source licensing. To maintain anonymity during the review process, we temporarily withhold our trained models and training logs to avoid disclosing author information. Upon acceptance, we will release all materials on the Hugging Face platform. We encourage the AI community to build upon our work while maintaining strict ethical standards, with the goal of making AI alignment more accessible while serving human values and social good.

\section*{Acknowledgments}

This work was supported by the Innovation and Technology Commission - Mainland-Hong Kong Joint Funding Scheme (Grant No. MHP/038/23).

\section*{Contribution Statement}

Yang Zhang and Yu Yu contribute to this work equally.

\bibliographystyle{named}
\bibliography{ijcai25}

\clearpage
\appendix

\begin{appendices}





\section{Experiments Setup}\label{app:setup}

\subsection{Parameter Setting}\label{app:paras}

The parameter settings for the presented algorithms are presented in Tables~\ref{tab:setting}.

\begin{table}[h]
\centering
\caption{Parameter Settings for MARA}
\label{tab:setting}
\scalebox{0.95}{
\begin{tabular}{lr}
\hline
\textbf{Parameters} & \textbf{Values} \\
\hline
Training episodes & 20000 \\
Number of trajectories collection workers & 7\\ 
Batch size & 1024 \\
Learning rate of actor network & 0.0003 \\
Learning rate of critic network & 0.0003 \\
Learning rate of entropy coefficient & 0.0003 \\
KL divergence coefficient $\lambda$ & 0.1\\
Initial entropy coefficient & 0.8\\
Discount factor & 0.99\\
Buffer capacity & 1e6\\
Network layers & 3 \\
Hidden size & [4096, 1024, 256] \\
Target network update rate & 0.005 \\
Max response length limit & 512\\
p of Top-p & 0.95\\
k of top-k & 50\\
\hline
\end{tabular}}
\end{table}

\subsection{Evaluation Metrics}\label{app:metrics}

To systematically compare different alignment approaches, we utilize the scores from reward models: beaver-7b-v1.0-reward for helpfulness evaluation and beaver-7b-v1.0-cost for harmlessness assessment. The comparison methodology operates as follows:

\begin{itemize}
\item A win ($N_w$ +1) is recorded when an alignment approach achieves higher scores in both helpfulness and harmlessness dimensions
\item A lose ($N_l$ +1) is registered when an approach scores lower in both dimensions
\item A tie ($N_e$ +1) is counted in all other scenarios where the superiority is not conclusive
\end{itemize}

The rationale behind this comparison methodology stems from our observation that many aligned models simply refuse to respond to adversarial prompts. While such behavior ensures safety, it compromises utility. Therefore, we propose a more rigorous metric where a response is considered superior only when it demonstrates both enhanced helpfulness and harmlessness simultaneously.

\section{More Experiment Results}\label{app:results}

\subsection{Detailed Results about Table~\ref{tab:main_1} - \ref{tab:main_2}}\label{app:details}

Table~\ref{tab:details_1} presents the detailed results about Table~\ref{tab:main_1} in terms of the win, tie,
and lose number when applying MARA compared to using the original LLM alone. For evaluation, each entry presents three numbers: the number of times MARA outperforms the original LLM (win), the number of times they perform equally well (tie), and the number of times MARA underperforms (lose). For example, when applying MARA to Llama 3-8B on the SafeRLHF dataset, it achieves better performance in 93 cases, equal performance in 80 cases, and worse performance in 26 cases compared to using Llama 3-8B alone.

Similarly, Table~\ref{tab:details_2} - \ref{tab:details_4} present the detailed results about Table~\ref{tab:main_2} in terms of the win, tie, and lose number when comparing MARA to other alignment approaches in PKU-SafeRLHF, BeaverTails, and HarmfulQA datasets, respectively. For instance, when comparing MARA with RLHF on the BeaverTails dataset using Llama 3-8B, MARA achieves better performance in 251 cases, equal performance in 353 cases, and worse performance in 96 cases. These comprehensive statistics across different datasets and baseline methods demonstrate the consistent effectiveness of our proposed MARA approach across various model scales and alignment baselines.


\begin{table*}[t]
\centering
\caption{Performance improvements of MARA across PKU-SafeRLHF, BeaverTails, and HarmfulQA datasets. Each entry shows the win, tie, or lose number when applying MARA compared to using the original LLM alone.}
\scalebox{1}{
\begin{tabular}{l*{9}{c}}
\toprule
& \multicolumn{9}{c}{Upstream LLM + MARA \textit{vs.} Upstream LLM} \\
\cmidrule(lr){2-10}
& \multicolumn{3}{c}{SafeRLHF} & \multicolumn{3}{c}{BeaverTails} & \multicolumn{3}{c}{HarmfulQA} \\
\cmidrule(lr){2-4} \cmidrule(lr){5-7} \cmidrule(lr){8-10}
Upstream LLM & Win & Tie & Lose & Win & Tie & Lose & Win & Tie & Lose \\
\midrule
Llama 3-8B & 93 & 80 & 26 & 355 & 248 & 97 & 296 & 150 & 42 \\
Llama 3.1-8B & 98 & 82 & 19 & 335 & 278 & 87 & 331 & 129 & 28 \\
Llama 3.2-1B & 104 & 81 & 14 & 334 & 254 & 112 & 329 & 120 & 39 \\
Llama 3.2-3B & 98 & 68 & 33 & 294 & 318 & 88 & 236 & 189 & 63 \\
Mistral-7B-v0.1 & 44 & 133 & 22 & 200 & 422 & 78 & 93 & 353 & 42 \\
Mistral-7B-v0.2 & 43 & 122 & 34 & 200 & 400 & 100 & 100 & 312 & 76 \\
Mistral-7B-v0.3 & 59 & 115 & 25 & 206 & 400 & 94 & 97 & 348 & 43 \\
\bottomrule
\end{tabular}}
\label{tab:details_1}
\end{table*}


\begin{table*}[t]
\centering
\caption{Performance comparison of MARA against RLHF, DPO, and \textit{Aligner} on PKU-SafeRLHF dateset. Each entry shows the win, tie, or lose number when applying MARA compared to other alignment methods.}
\scalebox{1}{
\begin{tabular}{l*{9}{c}}
\toprule
& \multicolumn{3}{c}{MARA \textit{vs.} RLHF} & \multicolumn{3}{c}{MARA \textit{vs.} DPO} & \multicolumn{3}{c}{MARA \textit{vs.} \textit{Aligner}} \\
\cmidrule(lr){2-4} \cmidrule(lr){5-7} \cmidrule(lr){8-10}
Upstream LLM & Win & Tie & Lose & Win & Tie & Lose & Win & Tie & Lose \\
\midrule
Llama 3-8B & 81 & 96 & 22 & 65 & 99 & 35 & 52 & 106 & 41 \\
Llama 3.1-8B & 68 & 106 & 25 & 43 & 113 & 43 & 38 & 124 & 37 \\
Llama 3.2-1B & 89 & 99 & 11 & 47 & 106 & 46 & 27 & 133 & 39 \\
Llama 3.2-3B & 50 & 116 & 33 & 54 & 93 & 52 & 37 & 121 & 41 \\
Mistral-7B-v0.1 & 36 & 143 & 20 & 49 & 120 & 30 & 45 & 130 & 24 \\
Mistral-7B-v0.2 & 66 & 116 & 17 & 68 & 108 & 23 & 56 & 116 & 27 \\
Mistral-7B-v0.3 & 49 & 129 & 21 & 60 & 113 & 26 & 53 & 125 & 21 \\
\bottomrule
\end{tabular}}
\label{tab:details_2}
\end{table*}


\begin{table*}[t]
\centering
\caption{Performance comparison of MARA against RLHF, DPO, and \textit{Aligner} on BeaverTails dateset. Each entry shows the win, tie, or lose number when applying MARA compared to other alignment methods.}
\scalebox{1}{
\begin{tabular}{l*{9}{c}}
\toprule
& \multicolumn{3}{c}{MARA \textit{vs.} RLHF} & \multicolumn{3}{c}{MARA \textit{vs.} DPO} & \multicolumn{3}{c}{MARA \textit{vs.} \textit{Aligner}} \\
\cmidrule(lr){2-4} \cmidrule(lr){5-7} \cmidrule(lr){8-10}
Upstream LLM & Win & Tie & Lose & Win & Tie & Lose & Win & Tie & Lose \\
\midrule
Llama 3-8B & 251 & 353 & 96 & 254 & 296 & 150 & 172 & 375 & 153 \\
Llama 3.1-8B & 259 & 378 & 63 & 202 & 398 & 100 & 120 & 438 & 142 \\
Llama 3.2-1B & 228 & 400 & 72 & 181 & 339 & 180 & 97 & 422 & 181 \\
Llama 3.2-3B & 189 & 374 & 137 & 179 & 361 & 160 & 120 & 413 & 167 \\
Mistral-7B-v0.1 & 127 & 445 & 128 & 221 & 379 & 100 & 195 & 429 & 76 \\
Mistral-7B-v0.2 & 191 & 436 & 73 & 263 & 363 & 74 & 197 & 400 & 103 \\
Mistral-7B-v0.3 & 191 & 436 & 73 & 278 & 353 & 69 & 226 & 383 & 91 \\
\bottomrule
\end{tabular}}
\label{tab:details_3}
\end{table*}


\begin{table*}[t]
\centering
\caption{Performance comparison of MARA against RLHF, DPO, and \textit{Aligner} on HarmfulQA dateset. Each entry shows the win, tie, or lose number when applying MARA compared to other alignment methods.}
\scalebox{1}{
\begin{tabular}{l*{9}{c}}
\toprule
& \multicolumn{3}{c}{MARA \textit{vs.} RLHF} & \multicolumn{3}{c}{MARA \textit{vs.} DPO} & \multicolumn{3}{c}{MARA \textit{vs.} \textit{Aligner}} \\
\cmidrule(lr){2-4} \cmidrule(lr){5-7} \cmidrule(lr){8-10}
Upstream LLM & Win & Tie & Lose & Win & Tie & Lose & Win & Tie & Lose \\
\midrule
Llama 3-8B & 238 & 232 & 18 & 219 & 188 & 81 & 105 & 306 & 77 \\
Llama 3.1-8B & 121 & 274 & 93 & 113 & 270 & 105 & 76 & 278 & 134 \\
Llama 3.2-1B & 233 & 212 & 43 & 173 & 215 & 100 & 91 & 259 & 138 \\
Llama 3.2-3B & 93 & 297 & 98 & 110 & 244 & 134 & 73 & 279 & 136 \\
Mistral-7B-v0.1 & 58 & 372 & 58 & 96 & 326 & 66 & 66 & 359 & 63 \\
Mistral-7B-v0.2 & 113 & 345 & 30 & 156 & 283 & 49 & 110 & 289 & 89 \\
Mistral-7B-v0.3 & 144 & 330 & 14 & 163 & 279 & 46 & 108 & 328 & 52 \\
\bottomrule
\end{tabular}}
\label{tab:details_4}
\end{table*}

\subsection{The Ablation on the Size of Candidate Token Set.}\label{app:size}

\begin{table*}[t]
\centering
\caption{Experimental analysis of how token candidate length affects model alignment performance through Top-k and Top-p sampling strategies on PKU-SafeRLHF dataset. The results measure the performance improvements (in percentage) over the upstream model in terms of preference rate.}
\scalebox{0.85}{
\begin{tabular}{l l *{9}{r}}
\toprule
& & \multicolumn{3}{c}{Top k=10} & \multicolumn{3}{c}{Top k=20} & \multicolumn{3}{c}{Top k=40} \\
\cmidrule(lr){3-5} \cmidrule(lr){6-8} \cmidrule(lr){9-11}
Upstream LLM & Metrics & p=0.7 & p=0.8 & p=0.9 & p=0.7 & p=0.8 & p=0.9 & p=0.7 & p=0.8 & p=0.9 \\
\midrule
\multirow{2}{*}{Llama-3.1-8B}
& Length & 6 & 7 & 8 & 10 & 12 & 14 & 14 & 19 & 27  \\
& Preference rate & +26.63\% & +28.64\% & +40.70\% & +28.64\% & +28.64\% & +45.23\% & +28.64\% & +30.65\% & +44.22\% \\
\midrule
\multirow{2}{*}{Llama-3.2-1B}
& Length & 6 & 7 & 9  & 10 & 13 & 16  & 16 & 20 & 30  \\
& Preference rate & +21.11\% & +19.60\% & +34.17\% & +25.13\% & +30.15\% & +37.69\% & +20.60\% & +34.17\% & +33.67\% \\
\bottomrule
\end{tabular}}
\label{tab:abla_size}
\end{table*}

To investigate how the candidate token set size influences model performance, we conducted ablation experiments with different Top-p and Top-k configurations using Llama-3.1-8B and Llama-3.2-1B. Table~\ref{tab:abla_size} shows that increasing p and k values leads to larger candidate token sets, with lengths growing from 6 to 27 tokens for Llama-3.1-8B and 6 to 30 tokens for Llama-3.2-1B at k=40. Notably, larger candidate sets generally yield better performance: Llama-3.1-8B's preference rate improvement increases from +26.63\% (k=10, p=0.7) to +44.22\% (k=40, p=0.9), while Llama-3.2-1B's improvement rises from +21.11\% (k=10, p=0.7) to +33.67\% (k=40, p=0.9) as the candidate set expands. This enhancement can be attributed to the increased diversity in token selection enabled by larger candidate sets.

\subsection{The Ablation on KL Divergence Coefficients}\label{app:KL}

\begin{figure*}[t]
    \centering
    \begin{subfigure}[b]{0.495\textwidth}
        \centering
        \includegraphics[width=\textwidth]{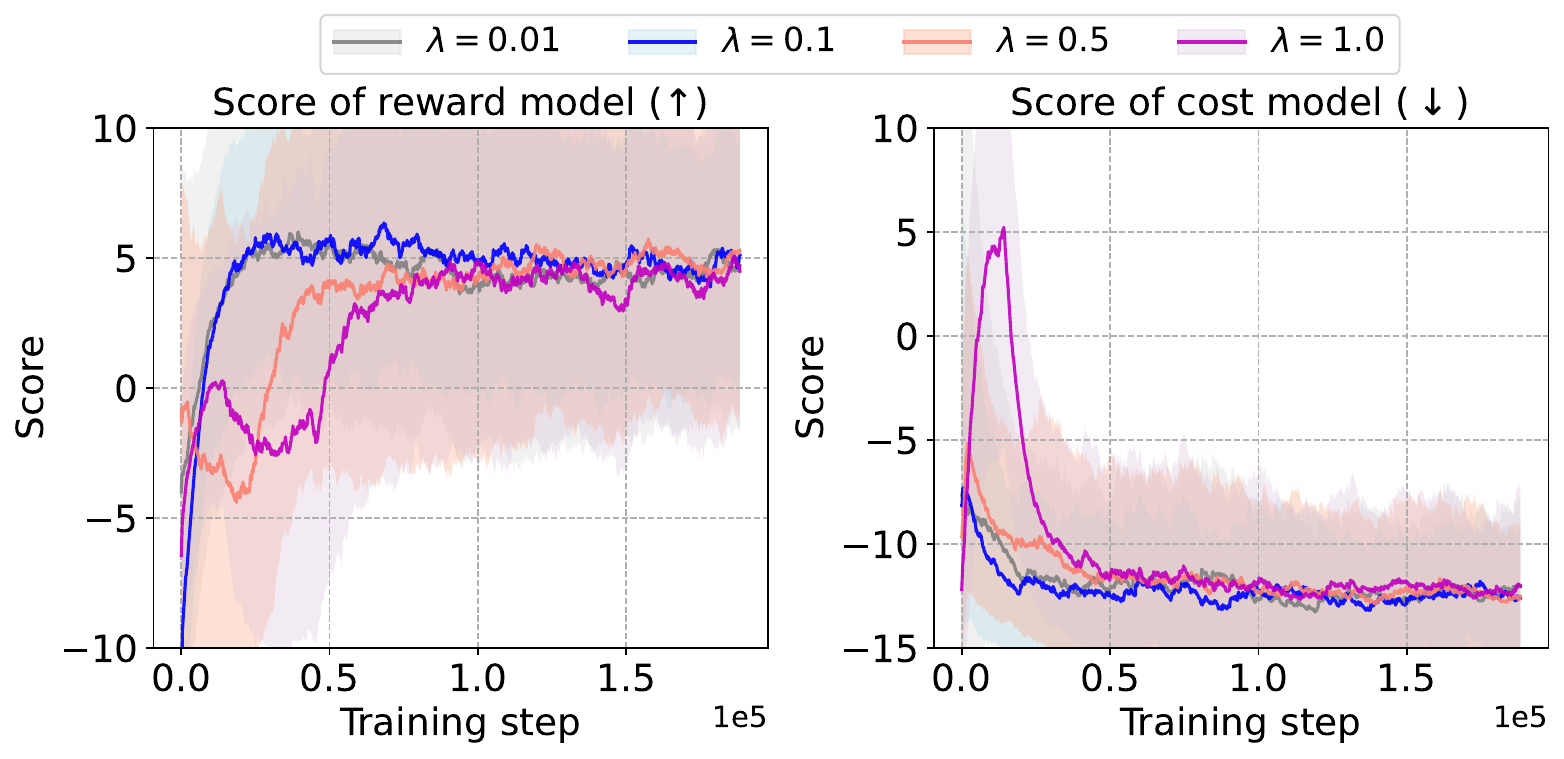}
        \caption{Llama-3.2-1B}
        \label{fig:ablation_lambda_llama}
    \end{subfigure}
    \hfill
    \begin{subfigure}[b]{0.495\textwidth}
        \centering
        \includegraphics[width=\textwidth]{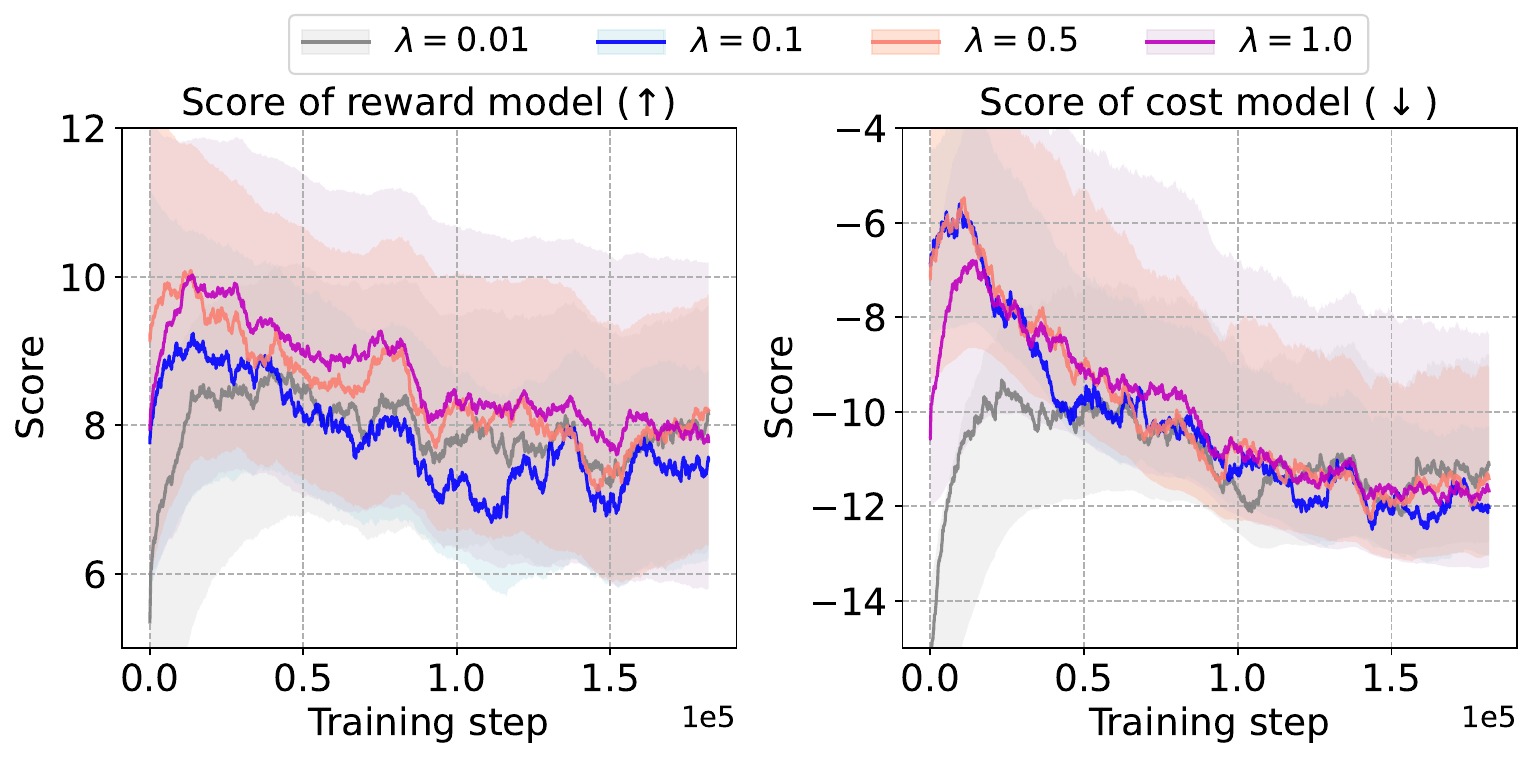}
        \caption{Mistral-7B-v0.3}
        \label{fig:ablation_lambda_mistral}
    \end{subfigure}
    \caption{Ablation experiments for the selection of KL divergence coefficient $\lambda$ on the algorithm performance. The experiments are performed on PKU-SafeRLHF dataset with different upstream models.}
    \label{fig:ablation_lambda}
\end{figure*}

To investigate the influence of KL divergence coefficient $\lambda$ on model performance, we conduct experiments with different $\lambda$ values (0.01, 0.1, 0.5, and 1.0) using Llama-3.2-1B and Mistral-7B-v0.3 models. As shown in Figure~\ref{fig:ablation_lambda}, we track both reward model scores and cost model scores during training. For Llama-3.2-1B (Figure~\ref{fig:ablation_lambda}a), while different $\lambda$ values lead to varying training dynamics initially, all configurations eventually converge to similar reward scores around 5 and cost scores around -10. Similarly, Mistral-7B-v0.3 (Figure~\ref{fig:ablation_lambda}b) exhibits convergence behavior where different $\lambda$ values ultimately reach comparable performance levels, with reward scores stabilizing around 8 and cost scores around -12. This convergence phenomenon suggests that the final model performance is relatively robust to the choice of $\lambda$, though the training stability and convergence speed differ. Based on the training dynamics and stability considerations, we choose $\lambda=0.1$ as our default setting, which offers a good balance between convergence speed and training stability.

\subsection{Visualisation Resuls}\label{app:vis}

\begin{figure*}[t]
    \centering
    \includegraphics[width=\textwidth]{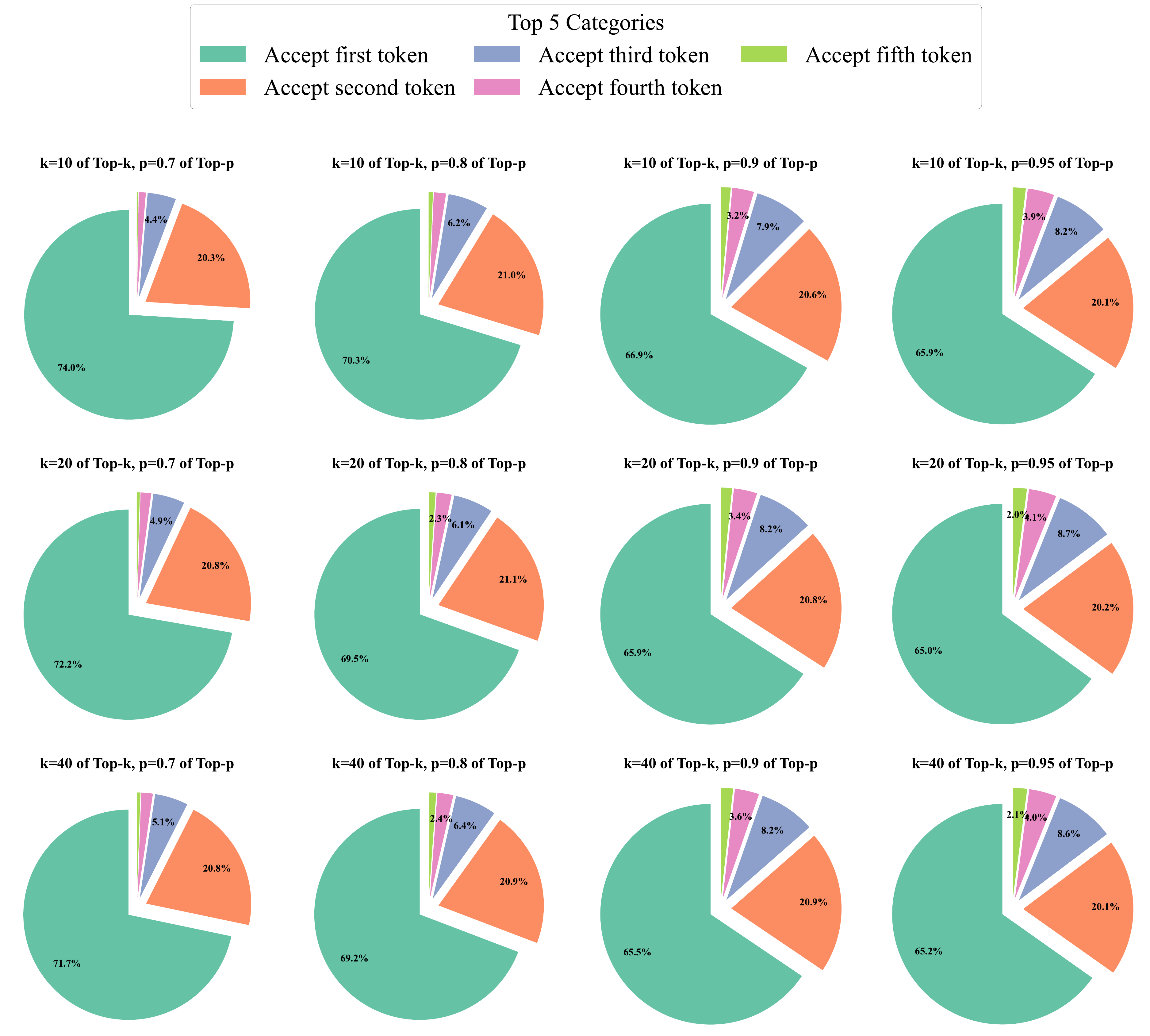}
    \caption{Distribution of accepted tokens across the candidate token set on PKU-SafeRLHF dataset with Llama-3.2-1B as the upstream LLM. Token positions with acceptance probabilities below 1\% are omitted for clarity.}
    \label{fig:token_distribution_2}
\end{figure*}

\begin{figure*}[t]
    \centering
    \includegraphics[width=\textwidth]{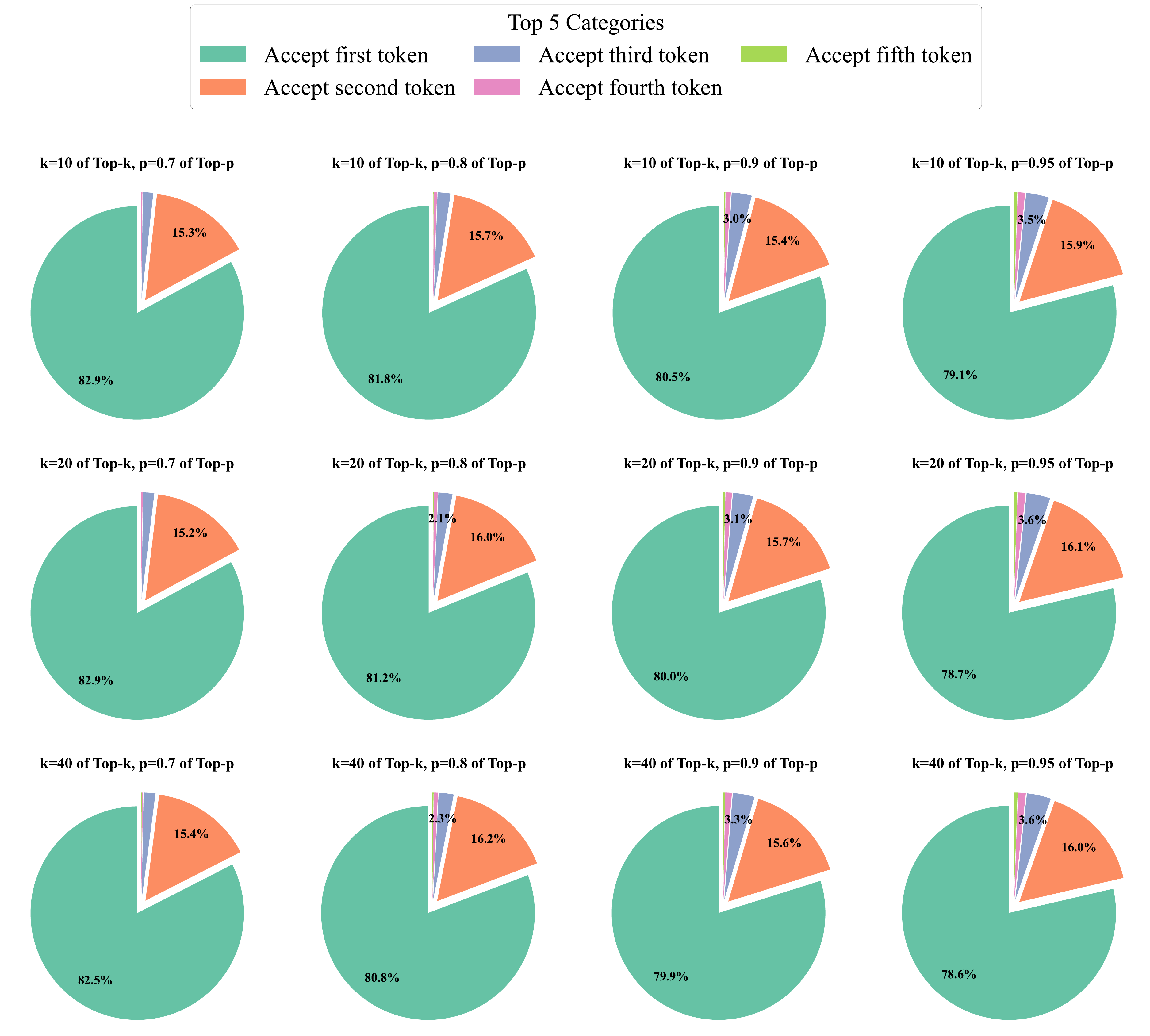}
    \caption{Distribution of accepted tokens across the candidate token set on PKU-SafeRLHF dataset with Llama-3.1-8B as the upstream LLM. Token positions with acceptance probabilities below 1\% are omitted for clarity.}
    \label{fig:token_distribution_3}
\end{figure*}

\begin{figure*}[t]
    \centering
    \includegraphics[width=\textwidth]{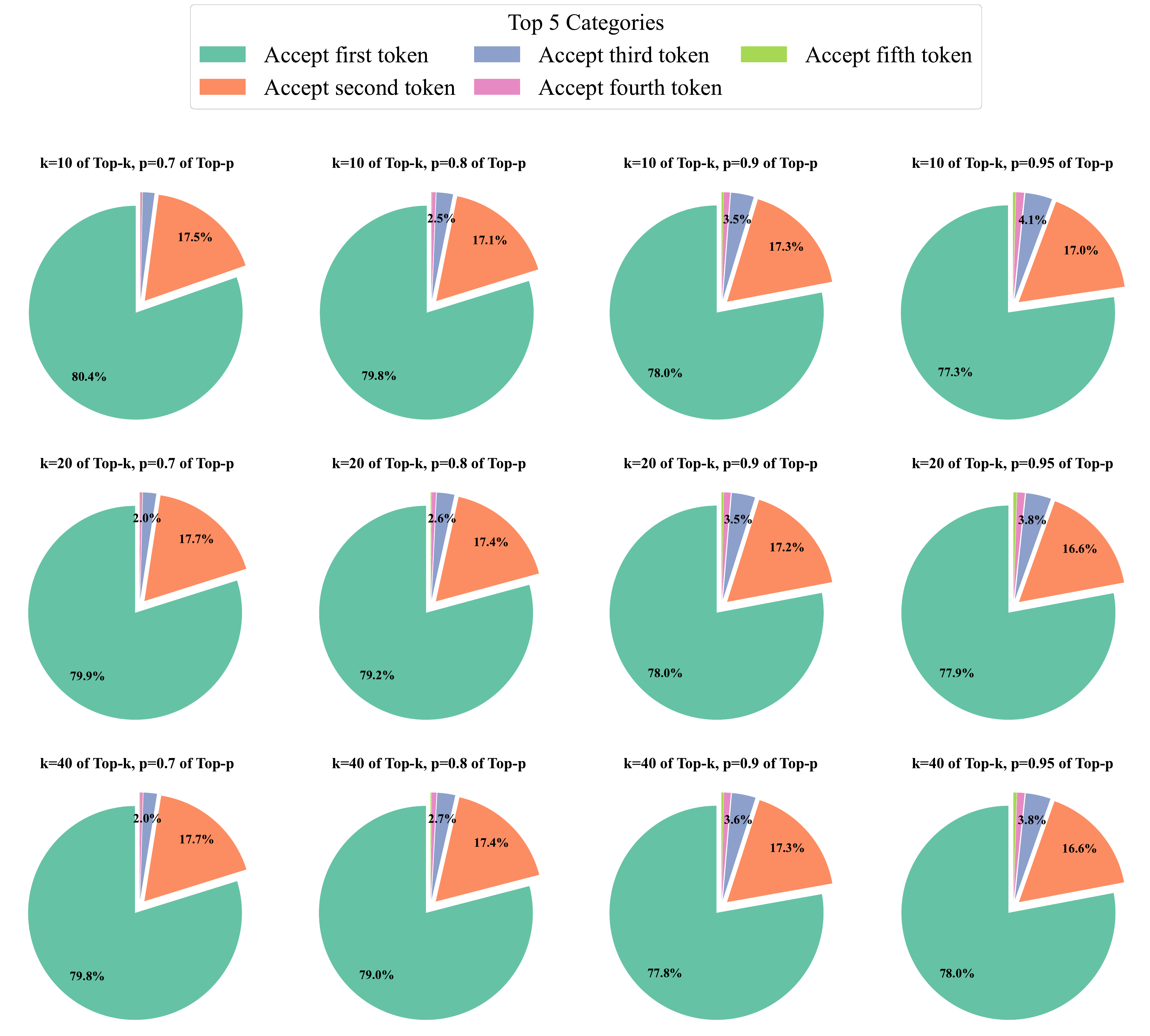}
    \caption{Distribution of accepted tokens across the candidate token set on PKU-SafeRLHF dataset with Mistral-7B-v0.3 as the upstream LLM. Token positions with acceptance probabilities below 1\% are omitted for clarity.}
    \label{fig:token_distribution_4}
\end{figure*}

Figures~\ref{fig:token_distribution_2} - \ref{fig:token_distribution_4} present the complete visualization results of the distribution of accepted tokens across the candidate token set on PKU-SafeRLHF dataset with Llama-3.2-1B, Llama-3.1-8B, and Mistral-7B-v0.3, respectively. From the complete results, the three same findings derived by Figure~\ref{fig:token_distribution} (dominance of first token, parameter sensitivity, and optimal decision window) can still be observed from Figures~\ref{fig:token_distribution_2} - \ref{fig:token_distribution_4}.

These patterns remain consistent across different model architectures and scales, suggesting they are inherent characteristics of the MARA approach rather than model-specific phenomena.

\end{appendices}

\end{document}